\documentclass[11pt, a4paper, copyright]{simular}

\usepackage{hyperref}
\usepackage{url}
\usepackage{wrapfig}
\usepackage{algorithm}
\usepackage{algpseudocode}
\usepackage{amsmath}
\usepackage{amsthm}
\usepackage{amssymb}
\usepackage{booktabs}
\usepackage{colortbl}
\usepackage{caption}
\usepackage{cleveref}
\usepackage{enumitem}
\usepackage{graphicx}
\usepackage{hyperref}
\usepackage{latexsym}
\usepackage{lipsum}
\usepackage{mathtools}
\usepackage{microtype}
\usepackage{multirow}
\usepackage{stmaryrd}
\usepackage{subfigure}
\usepackage{symbols}
\usepackage{tcolorbox}
\usepackage{url}
\usepackage{epigraph}      
\usepackage{xcolor}

\usepackage{mwe}
\usepackage{wrapstuff}
\definecolor{darkgreen}{RGB}{0, 100, 0}
\definecolor{apifn}{RGB}{0,80,120}

\NewDocumentEnvironment{rightwrapfig}{O{0.4\columnwidth}}{%
  \begin{wrapstuff}[r,width=#1]
    \centering
}{%
  \end{wrapstuff}%
}
\makeatletter
\let\cp@subsection\subsection
\let\cp@subsubsection\subsubsection
\renewcommand{\subsection}{\wrapstuffclear\cp@subsection}
\renewcommand{\subsubsection}{\wrapstuffclear\cp@subsubsection}
\makeatother

\newcommand{\passk}[1]{{Pass\textasciicircum#1}}

\newif\ifcomments
\commentstrue   

\ifcomments
  \newcommand{\hyewon}[1]{{\color{orange} Hyewon: #1}} 
  \newcommand{\jc}[1]{{\color{red} jc: #1}}
  \newcommand{\eric}[1]{{\color{blue} Eric: #1}}
  \newcommand{\richard}[1]{{\color{darkgreen} Richard: #1}}
  \newcommand{\ang}[1]{{\color{magenta} Ang: #1}}
  \newcommand{\minh}[1]{{\color{magenta} Minh: #1}}
\else
  \newcommand{\hyewon}[1]{}
  \newcommand{\jc}[1]{}
  \newcommand{\eric}[1]{}
  \newcommand{\richard}[1]{}
  \newcommand{\ang}[1]{}
  \newcommand{\minh}[1]{}
\fi
\newcommand{\framework}{Neuro-Symbolic Policy Iteration}
\newcommand{\frameworkabbr}{NSPI}

\providecommand{\passk}[1]{{Pass\textasciicircum#1}}
\providecommand{\framework}{Neuro-Symbolic Policy Iteration}
\providecommand{\frameworkabbr}{NSPI}
\providecommand{\Gcal}{\mathcal{G}}
\providecommand{\Ebb}{\mathbb{E}}
\providecommand{\Pbb}{\mathbb{P}}
\providecommand{\Acal}{\mathcal{A}}
\providecommand{\Wcal}{\mathcal{W}}
\providecommand{\defeq}{\coloneqq}

\usepackage{booktabs}
\usepackage{amsmath}
\usepackage{amssymb}
\usepackage{mathtools}
\usepackage{graphicx}
\usepackage{xcolor}
\definecolor{apifn}{RGB}{0,80,120}
\usepackage{algorithm}
\usepackage{algpseudocode}
\usepackage{cleveref}
\usepackage{multirow}
\usepackage{epigraph}
\usepackage{caption}

\title{Neuro-Symbolic Computer Use: Learning Reusable Policies for Reliable and Efficient Execution}

\author{
Hyewon Suh\textsuperscript{$\dagger$},
Thanh Minh Nguyen,
Chih-Lun Lee,
Darrow Hartman\textsuperscript{$\ddagger$},
Lizhao Liu,
Xin Eric Wang,
Ang Li,
Jiachen Yang
}

\affiliations{\textsuperscript{$\dagger$}Georgia Institute of Technology, \textsuperscript{$\ddagger$}Stanford University; work done at Simular.\ }
\correspondingauthor{{\{hyewon, jiachen\}}@simular.ai}

\usepackage[authoryear, sort&compress, round]{natbib}
\uselogo{}
\renewcommand{\today}{}

\begin{abstract}
Many computer tasks recur: the same workflow runs many times, with new inputs and from different starting states. Current computer-use agents re-plan every step of every run, which makes them costly and unreliable on such tasks. We introduce \emph{neuro-symbolic computer use}, in which a recurring workflow is executed by a learned policy rather than re-derived by an agent on each run. The policy fixes the decisions that are stable across runs (ordering, variables, loops, and branches) in executable code, and delegates observation-dependent decisions, such as grounding and state checks, to neural models. We learn these policies with \emph{neuro-symbolic policy iteration}: starting from one agent trajectory, it executes the policy, diagnoses failures with task-completion and step-level judges, and revises the code with a coding model informed by an agent's continuation from the point of failure, without access to the benchmark evaluator. Iterating on generated parameter and initial-state variants makes the policy reusable, and a pre-action verifier guards each state-mutating step at deployment. On OSWorld-Verified and ScienceBoard, the learned policies achieve the highest \passk{3} of all methods in all four settings, 3.6--15.8 points above the base agent, while cutting per-run cost by 15--217$\times$ and latency by 3.4--5.1$\times$. On OSWorld-Verified, policies built only on variants transfer to the held-out original tasks, exceeding AutoRPA by 8.6--17.5 points in \passk{3}.
\end{abstract}

\begin{document}
\maketitle

\epigraph{\itshape Civilization advances by extending the number of important operations which we can perform without thinking about them.}{--- Alfred North Whitehead}

\section{Introduction}
\label{sec:intro}

A person who files expense reports a thousand times does not rediscover the same procedure on the thousandth attempt. However, contemporary computer-use agents still do: they plan step-by-step on every run, as if it were the first \citep{xie2024osworld,agashe2025agent,agashe2025agents2,sun2025scienceboard,agents3}. For economically-valuable recurring workflows \citep{patwardhan2026gdpval} where the same algorithmic procedure only needs limited decision-making, agents that plan from scratch on every step of every run face two critical problems: unreliability over repeated runs of the same task \citep{gonzalez2026reliability}, and high cost and latency of frontier-model calls on long workflows \citep{yuan2026osworld2}. How can an agent turn execution experience into a reusable procedure that stays reliable and efficient as inputs and states vary?

Existing ways to automate such workflows sit at two extremes. At the neural extreme, computer-use agents decide every action from the current observation~\citep{agashe2025agents2,agents3,uitars,opencua}. They are flexible, but re-decide steps that never change, so every run is costly and lacks reliability. At the symbolic extreme, robotic process automation (RPA) runs a recorded or hand-written script \citep{van2018robotic}, which is cheap to repeat but breaks when a states changes and does not generalize beyond anticipated cases.

To solve this dilemma, we introduce the paradigm of \textbf{Neuro-Symbolic Computer Use}, in which a recurring workflow is executed by a learned neuro-symbolic policy instead of being re-derived by an agent on every run. Our core insight is that recurring workflows contain a mixture of deterministic transitions and state-dependent decisions, and that neuro-symbolic programs are reliable and efficient \emph{representations} of such workflows. In expense reporting for example, form fields almost never change, but the location of a ``submit'' button may vary and a check for login may be needed. We therefore represent the procedure as a neuro-symbolic policy: code fixes the stable decisions (algorithm, variables, and control flow), while neural primitives make state-dependent decisions at run time.

\begin{figure}[t]
\centering
\includegraphics[width=\linewidth]{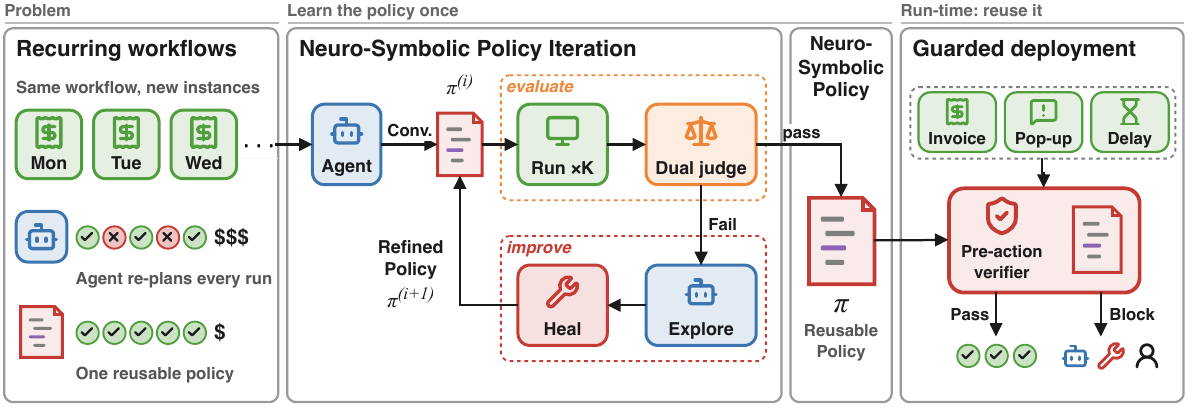}
\caption{
\textit{Recurring} computer workflows require reliable repeated execution with generalizability at low cost. In \textit{Neuro-Symbolic Policy Iteration}: a baseline agent's exploration seeds an initial neuro-symbolic policy, which preserves the workflow's stable structure in code while retaining neural primitives for observation-dependent decisions. Our framework iteratively executes, evaluates, diagnoses, and improves the policy with additional agent continuation. The resulting policy is reused across workflows and guarded with pre-action verifiers during run-time deployment.
}
\vspace{-10pt}
\label{fig:algorithm}
\end{figure}

To learn such policies, we introduce \textbf{\framework{} (\frameworkabbr)}, which synthesizes an initial policy from a base neural agent's demonstration trajectories and refines it by iterating between policy evaluation and policy improvement without ground truth rewards (\Cref{fig:algorithm}). A task-completion judge checks the end state of each iteration's policy rollout, and a step-level judge locates the first step that went wrong. Upon a failure verdict, an agent continues the task from the failed state, and a coding model revises the policy using this continuation and the judges' diagnosis. This execute--diagnose--explore--revise loop repeats until every rollout in an iteration passes or the budget runs out. At convergence, the policy itself represents a summary of the learned experience, and only here does it receive feedback from the benchmark evaluator. To test generalizability we also refine it on task variants, before its deployment along with a pre-action verifier guarding each state-mutating action against environment drift.

We evaluate \frameworkabbr{} on three properties that recurring workflows require: reliability across repeated runs, reuse on similar task instances, and efficiency. On OSWorld-Verified (henceforth OSWorld)~\citep{xie2024osworld} and ScienceBoard~\citep{sun2025scienceboard} with GPT-5.6 Terra and Claude Opus~5 as base agent models, the learned policies achieve the highest \passk{3} (the share of tasks solved in all three runs) of all methods in all four settings, 4.5--13.9 points above the base agent. We also learn policies only on synthetic OSWorld variants that change a task's parameters or its starting state, and evaluate them on the held-out original tasks, where they exceed AutoRPA~\citep{chen2026autorpa} on the same task families by 8.6-17.5 points in \passk{3}. Because a learned policy does not re-derive the workflow, it runs 15--217$\times$ cheaper and 3.4--5.1$\times$ faster than the base agent, and the one-time cost of learning it is recovered after 3--27 runs.

\section{Neuro-Symbolic Computer Use}
\label{sec:nscu}

\subsection{Problem statement: recurring workflows and reusable policies}
\label{sec:method:objective}

We consider recurring parametric computer tasks on open-world desktop environments, a common setting in personal logistics and business processes. A task $g$ is a natural-language instruction to be completed in a digital environment (a desktop application or a website) with transition function $T$ and initial state distribution $\rho$. Executing a policy $\pi$ from an initial state $s_0 \sim \rho$ produces a trajectory $\tau = (o_0, a_0, o_1, \dotsc, o_{|\tau|})$ of observations and actions. A held-out success indicator $R(\tau; g) \in \lbrace 0, 1 \rbrace$ (e.g., the evaluation script of a benchmark or a user's unknown preferences) determines whether $\tau$ completed $g$. We require a policy for each $g$ to succeed across a parameterized family $\Gcal$ of task instances similar to $g$ (e.g., ``book a flight on \{date\}'') and across initial states $s_0 \sim \rho$.

\textbf{Objective and reliability metric.} We seek a policy that maximizes reliability over the task family:
\begin{align}
\label{eq:objective}
\max_\pi \; \Ebb_{g \sim \Gcal} \big[\text{\passk{k}}(\pi; g)\big] ,
\end{align}
where \passk{k}~\citep{yao2024tau} is the probability that all $k$ independent executions of $\pi$ on $g$ succeed:
\begin{align}
\text{\passk{k}}(\pi; g) &\defeq \Pbb\big[R(\tau_1; g) = \dotsb = R(\tau_k; g) = 1\big] = p_\pi(g)^k , \label{eq:passk} \\
p_\pi(g) &\defeq \Ebb_{s_0 \sim \rho,\, \tau \sim (\pi, T)} \big[R(\tau; g)\big] , \nonumber
\end{align}
Given $n$ executions with $c$ successes, $\binom{c}{k} / \binom{n}{k}$ is the unbiased estimator of \passk{k} for any $k \leq n$.

\textbf{Amortized cost metric.} Let $C_\pi$ be the one-time cost (in dollars or wall-clock time) of constructing $\pi$ and $c_\pi$ its cost per execution; the amortized cost per execution over $n$ executions is
\begin{align}
\label{eq:amortized}
\bar{c}_\pi(n) \defeq \frac{C_\pi}{n} + c_\pi .
\end{align}
A pure neural agent $\mu$ has $C_\mu = 0$ (ignoring the costs of model training) but a high $c_\mu$ if it calls an expensive LLM to plan every step. In general, a policy $\pi$ with $C_\pi > 0$ and a lower $c_\pi$ would have lower amortized cost once $n$ exceeds the break-even point $n^\ast = C_\pi / (c_\mu - c_\pi)$.

\subsection{Neuro-symbolic policies}
\label{sec:method:framework:ns-policy-representation}

\begin{figure}[t]
\centering
\includegraphics[width=\linewidth]{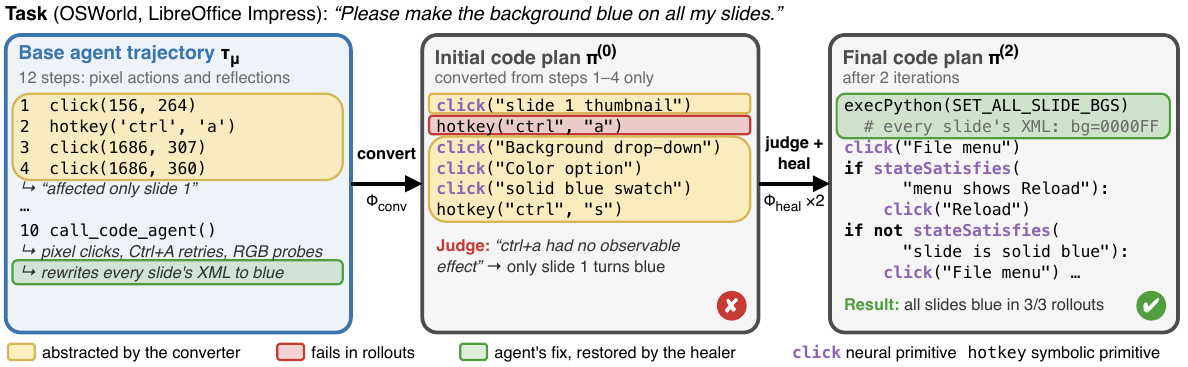}
\caption{
Example of \frameworkabbr{} converting agent trajectory into a reusable, abstracted neuro-symbolic policy and using judge feedback to identify and heal failures. The resulting policy combines state-dependent control flow, task parameters in code, and neural primitives for grounding 
(\Cref{sec:experiment_setup}).
}
\vspace{-10pt}
\label{fig:code_policy}
\end{figure}

A neuro-symbolic policy is a program $P$ in a scripting language whose statements are control flow (loops, conditionals, variable assignment, and recursion) and calls to primitives from a set $\Acal = \Acal_\text{sym} \cup \Acal_\text{neu}$ (detailed in \Cref{sec:appendix:action_space}). We write $\Pi$ for the class of policies defined by such programs. Symbolic primitives in $\Acal_\text{sym}$, such as keyboard shortcuts, shell commands, and file operations, execute without a model call. Neural primitives in $\Acal_\text{neu}$ call a model on the current observation $o_t$, and our work includes: 1) GUI grounding actions (e.g., \textcolor{apifn}{click(description)}) map natural-language element descriptions to screen coordinates using a visual grounding model; 2) a condition check $Q(\phi, o_t) \in \lbrace 0, 1 \rbrace$, e.g., \textcolor{apifn}{$\text{stateSatisfies(condition)}\rightarrow \lbrace 0, 1\rbrace$}, returns whether $o_t$ satisfies a natural-language condition $\phi$, and the control flow of $P$ can branch on its output. All neural primitives include the accessibility tree and the screenshot in $o_t$ as input by default, so the policy only needs to represent natural language inputs. \Cref{fig:code_policy} shows an example policy before and after refinement. At deployment, the program runtime applies each state-mutating primitive with the pre-action verifier of \Cref{sec:method:robustness:verifier}.

\section{\framework{}}
\label{sec:method}

\frameworkabbr{} first learns a policy that solves a task reliably (\Cref{sec:method:learning}) and then generalizes it beyond the instance it was learned on (\Cref{sec:method:reuse}); \Cref{fig:algorithm} shows both parts.

\subsection{Policy learning}
\label{sec:method:learning}
\label{sec:method:overview}

\subsubsection{Synthesis and iterative refinement}
\label{sec:method:framework:synthesis-and-refinement}
\label{sec:method:refinement}

\textbf{Policy iteration.} We search $\Pi$ for a policy that maximizes \Cref{eq:objective} by policy iteration~\citep{howard1960dynamic,sutton2018reinforcement}, which alternates between policy evaluation and policy improvement. A dual judge evaluates the policy (\Cref{sec:method:framework:policy-evaluation}), and a code healer improves it (\Cref{sec:method:framework:policy-improvement}). Together, the two judges are the analog of the critic in policy iteration. Despite not being fit to ground-truth rewards like a classical critic, they rely on the knowledge of the underlying LLMs to produce an evaluation of the current policy in natural language.

\textbf{Initialization.} Given a task $g$, a base agent $\mu$ produces a trajectory $\tau_\mu$, which need not be successful. A converter $\Phi_\text{conv}$ turns $\tau_\mu$ into the initial policy $\pi^{(0)}$: it concatenates the agent's actions into a single program, replaces pixel coordinates with natural-language descriptions of the target elements that the runtime grounds at execution time, and replaces values specific to $g$ with task parameters.

\textbf{Iteration.} At iteration $i$, we execute $\pi^{(i)}$ up to $K$ times, each from a freshly reset initial state, and judge each rollout $\tau^{(i)}_j$ with the dual judge of \Cref{sec:method:framework:policy-evaluation} (policy evaluation). If all $K$ rollouts pass the judge, we return $\pi^{(i)}$. Otherwise, we stop at the first failed rollout and produce $\pi^{(i+1)}$ with the code healer of \Cref{sec:method:framework:policy-improvement} (policy improvement). The loop runs for at most $N$ iterations and returns the last policy if no iteration passes. Note again that $R$ is not used; we apply it only the returned policy to measure \passk{k}. Appendix~\ref{sec:appendix:method:loop} gives further details of the loop.

\subsubsection{Policy evaluation with a dual judge}
\label{sec:method:framework:policy-evaluation}
\label{sec:method:diagnosis}

We evaluate each policy rollout with two judges to detect and localize failures and provide diagnosis.

\textbf{Task-completion judge.} The task-completion judge $J_\text{task}$ is a VLM agent that receives only the task $g$ (and the answer submitted by the policy, for question-answering tasks) and probes the end state of the machine with read-only tools, such as shell commands and file reads. It returns a verdict $v_\text{task} \in \lbrace \text{success}, \text{failure}, \text{abstain} \rbrace$ with the evidence $e_\text{task}$ it collected, and it abstains when success depends on transient states during task execution; an applicability gate skips it for tasks whose success cannot be checked from persistent state.

\textbf{Step-level judge.} The step-level judge runs on every rollout, in parallel with $J_\text{task}$, and has two stages. First, for each transition $t$, an observer VLM describes the change between consecutive observations, $\delta_t = J_\text{obs}(o_t, o_{t+1})$, without access to the task, the code, or the stated intent of the step, so that each description is based only on observable evidence. Second, a localizer VLM reads the task, the policy, and all observer outputs, and returns a verdict, the first step at which the rollout diverged from the task, and a named failure mode, $(v_\text{loc}, t^\ast, m) = J_\text{loc}(g, \pi, \delta_{0:|\tau|-1})$.

\textbf{Verdict and diagnostic.} A rollout passes only if $v_\text{loc}$ is success and $v_\text{task}$ is success or $J_\text{task}$ was skipped by the applicability gate, so an abstention by $J_\text{task}$ counts as a failure. An aggregator LLM composes the outputs of both judges, $(v_\text{task}, e_\text{task})$ and $(v_\text{loc}, t^\ast, m)$, into a diagnostic $d$ for the healer. We measure how often the judges agree with the benchmark evaluation scripts in Section~\ref{sec:results:judge}. \Cref{alg:diagnose} states the procedure, and Appendix~\ref{sec:appendix:judges} gives the details of both judges.

\subsubsection{Policy improvement with a code healer}
\label{sec:method:framework:policy-improvement}
\label{sec:method:improvement}

When rollout $\tau^{(i)}_j$ fails, the base agent $\mu$ continues the task from the end state of $\tau^{(i)}_j$, given $g$, the policy $\pi^{(i)}$, its execution log, and the diagnostic $d$, and produces a continuation $\tau_\text{cont}$. A summarizer LLM $\Phi_\text{sum}$ condenses it into a short description of the actions that made progress. The healer $\Phi_\text{heal}$, a coding LLM, then rewrites the policy:
\begin{align}
\label{eq:heal}
\pi^{(i+1)} = \Phi_\text{heal}\big(\pi^{(i)}, \tau^{(i)}_j, d, \Phi_\text{sum}(\tau_\text{cont}); g\big) ,
\end{align}
where $\tau^{(i)}_j$ enters as its execution log, error message, and a subset of its screenshots. Upon execution errors, we skip the judges and agent continuation, giving the error directly to the healer.

\subsection{Policy generalization}
\label{sec:method:reuse}
\label{sec:method:generalization}
\label{sec:method:robustness}

\frameworkabbr{} generalizes to anticipated task variations by running policy iteration on generated variants that perturb the task's parameters and initial state, so that the healer must write code that reads the parameters of the instance and handles each initial state (\Cref{sec:method:robustness:variations}). \Cref{sec:results:generalization} tests generalization to held-out tasks. Unanticipated environment nonstationarity are caught at deployment by a pre-action verifier, which can hand the task to a general agent or escalate it for human review in sensitive domains such as healthcare and finance (\Cref{sec:method:robustness:verifier}).

\subsubsection{Robust learning under workflow perturbations}
\label{sec:method:robustness:variations}

Given a target task $g$, we construct variants of $g$ in two categories of workflow variation, run policy iteration on the variants, and hold out $g$ itself for evaluation (\Cref{fig:variant_generation}). \textbf{Initial-state variants} keep the instruction of $g$ and start the environment in a different realistic state, e.g., the target document is closed, a dialog blocks the window, or the output file already holds stale values. We run the base agent on natural language instructions to set up such initial states. \textbf{Parameter variants} keep the initial state of $g$ and change the argument values in its instruction, e.g., a date, a count, a file name, or a setting value. An LLM proposes multiple variants per category, and a validator rejects any that changes the goal of the task or the content of taks setup files. Task variants are presented to \frameworkabbr{} sequentially, thereby exposing hard-coded values and unhandled states, forcing the policy to avoid overfitting any single variant. \Cref{sec:experiment_setup} describes the variants we construct for each benchmark.

\subsubsection{Pre-action verification}
\label{sec:method:robustness:verifier}
\label{sec:method:verifier}

We check the precondition \citep{fikes1971strips} of every state-mutating primitive using a fast and calibrated VLM classifier before deploying the final policy (\Cref{fig:verifier} in Appendix~\ref{sec:appendix:verifier}).

\textbf{Inputs and outputs.} The verifier $V_\theta$ receives the task, the primitive action (function call with inputs), a natural-language explanation of the step from the policy's in-line comments at (or before) the action, current observation, and the primitives already executed in this run (\Cref{eq:verifier_input} in Appendix~\ref{sec:appendix:verifier}), and it either passes or blocks the primitive. On a block, the particular application setting may choose among handing the task to a stateful agent to continue, running \frameworkabbr{}, or escalating to a human.

\textbf{Decision rule.} We prompt a VLM to answer with one token, yes if the action should run now and no otherwise, following a fixed rubric (Appendix~\ref{sec:appendix:verifier}). The verifier blocks when $p_\text{no}$, the renormalized probability mass of the no answers among the 20 most likely tokens, reaches a threshold $\theta$ set on a calibration set (\Cref{eq:pno} in Appendix~\ref{sec:appendix:verifier}). Using $p_\text{no}$ instead of the greedy answer turns the trade-off between wrong blocks and wrong passes into a single parameter $\theta$.

\section{Experiments and Results}

\subsection{Experimental setup}
\label{sec:experiment_setup}

\paragraph{Benchmarks}
We evaluate on OSWorld-Verified~\citep{xie2024osworld} and ScienceBoard~\citep{sun2025scienceboard}. We use 361 OSWorld tasks, excluding Google Drive tasks that require unavailable credentials, and 169 ScienceBoard tasks spanning six scientific desktop applications. Further evaluation settings are detailed in Appendix~\ref{sec:appendix:repeated-execution}. We also explore longer workflows on OSWorld-V2~\citep{yuan2026osworld2} as a separate extension (\Cref{sec:results:longer-tasks}).

\paragraph{Compared methods}
Using GPT-5.6 Terra~\citep{openai2026gpt56} and Claude Opus~5 \citep{claudeOpus5} as backbone models, we compare our method with three baselines. The \textbf{base agent} plans each action from the current observation and solves each task anew on every run, without retaining experience across runs. \textbf{Batch2Code} is a conversion-only variant of \frameworkabbr{} that uses the same code-policy without iterative refinement. It constructs the policy from multiple base-agent trajectories, giving it more initial trajectory evidence than our single-trajectory initialization. \textbf{ASI}~\citep{wang2025inducing} acquires reusable programmatic skills while retaining an agent to select and execute them. For reuse across parameter and initial-state variations, we compare with \textbf{AutoRPA}~\citep{chen2026autorpa}, adapted to learn reusable computer-use workflows from natural-language requests and execution experience, without benchmark-supplied task templates, parameter schemas, or oracle feedback. Further details on experiment setups and configurations are included in Appendix~\ref{sec:appendix:repeated-execution}.

\paragraph{Workflow variation}
We evaluate whether a learned policy transfers to a held-out instance of the same workflow. We constructed the workflow variations of the subset of OSWorld tasks. Parameter variants change request arguments, while initial-state variants preserve the request but alter the starting environment. We study 40 parameter and 35 initial-state families with up to three admitted variants. \frameworkabbr{} and AutoRPA both use GPT-5.6 Terra and a shared domain setup, without access to the benchmark evaluator. They process the same variants sequentially, refining a single policy as new instances expose task-specific assumptions. We then freeze the policy and evaluate \passk{3} on the held-out original task without further adaptation. We include implementation details in Appendix~\ref{sec:appendix:task-variants-baseline}.

\vspace{-5pt}
\paragraph{Metrics}
We report \passk{k} for $k\in\{1,2,3\}$, estimated from three evaluation trials per task as defined in \Cref{sec:method:objective}; \passk{3} measures success in all three trials. We average over the full task or family set, retaining construction failures in the denominator. We report per-execution cost and wall-clock latency separately from one-time construction cost and time, and analyze their amortization over repeated use (\Cref{sec:results:cost_amortization}).

\subsection{Reliability across repeated executions}

\begin{table}[t]
    \centering
    \small
    
\setlength{\tabcolsep}{4.5pt}%
\renewcommand{\arraystretch}{1.12}%
\definecolor{benchband}{gray}{0.88}%
\definecolor{oursrow}{RGB}{226,240,217}%
\newcommand{\ongoing}{\ifcomments\textsuperscript{\color{orange}$\circ$}\fi}%
\newcommand{\modelspan}[1]{\multirow{-4}{*}{\textit{#1}}}%
\ifdefined\mainrescol\else\newlength{\mainrescol}\fi
\setlength{\mainrescol}{\dimexpr(0.8\linewidth-0.95in-1.0in-10\tabcolsep)/4\relax}%
\begin{tabular}{@{}
    w{l}{0.95in}
    w{l}{1.0in}
    *{4}{w{c}{\mainrescol}}
    @{}}
    \toprule
    & & \multicolumn{2}{c}{\textbf{Success rate (\%)} $\uparrow$} & \multicolumn{2}{c}{\textbf{Per run} $\downarrow$} \\
    \cmidrule(lr){3-4}\cmidrule(lr){5-6}
    \textbf{Model} & \textbf{Method}
        & \textbf{\passk{1}}
        & \textbf{\passk{3}}
        & \textbf{Cost}
        & \textbf{Time} \\
    \midrule
        \rowcolor{benchband}
        \multicolumn{6}{c}{\rule{0pt}{2.4ex}\textbf{ScienceBoard} {\mdseries\scriptsize(169 tasks)}} \\
         & Base agent
            & \underline{27.0} & \underline{13.6} & 15$\times$ & 4.6$\times$ \\
         & Batch2Code
            & 15.4 & 11.8 & \underline{1.1}$\times$ & \underline{1.0}$\times$ \\
         & ASI
            & 24.3 & 11.8 & 16$\times$ & 5.1$\times$ \\
        \rowcolor{oursrow}
        \modelspan{GPT-5.6 Terra} & \textbf{\frameworkabbr{} (ours)}
            & \textbf{29.2} & \textbf{26.0} & \textbf{0.00536} & \textbf{40} \\
        \addlinespace[0.35em]
         & Base agent
            & 38.7 & \underline{29.6} & 65$\times$ & 3.9$\times$ \\
         & Batch2Code
            & 12.8 & 9.5 & \textbf{0.9}$\times$ & \textbf{0.9}$\times$ \\
         & ASI
            & \underline{39.8} & 26.0 & 68$\times$ & 3.8$\times$ \\
        \rowcolor{oursrow}
        \modelspan{Claude Opus 5} & \textbf{\frameworkabbr{} (ours)}
            & \textbf{48.7} & \textbf{42.6} & \underline{0.00790} & \underline{64} \\
        \midrule
        \rowcolor{benchband}
        \multicolumn{6}{c}{\rule{0pt}{2.4ex}\textbf{OSWorld} {\mdseries\scriptsize(361 tasks)}} \\
         & Base agent
            & \textbf{55.9} & 34.6 & 67$\times$ & 5.1$\times$ \\
         & Batch2Code
            & 51.5 & \underline{47.1} & \textbf{0.9}$\times$ & \textbf{0.8}$\times$ \\
         & ASI
            & 53.1 & 31.9 & 62$\times$ & 6.2$\times$ \\
        \rowcolor{oursrow}
        \modelspan{GPT-5.6 Terra} & \textbf{\frameworkabbr{} (ours)}
            & \underline{53.6} & \textbf{50.4} & \underline{0.00716} & \underline{56} \\
        \addlinespace[0.35em]
         & Base agent
            & \textbf{74.1} & \underline{68.4} & 217$\times$ & 3.4$\times$ \\
         & Batch2Code
            & 46.7 & 45.2 & \textbf{0.6}$\times$ & \textbf{0.5}$\times$ \\
         & ASI
            & \underline{73.8} & 67.6 & 381$\times$ & 4.2$\times$ \\
        \rowcolor{oursrow}
        \modelspan{Claude Opus 5} & \textbf{\frameworkabbr{} (ours)}
            & 72.9 & \textbf{72.0} & \underline{0.00965} & \underline{103} \\
    \bottomrule
\end{tabular}%

    \caption{
        Reliability across repeated executions on ScienceBoard and OSWorld. Per-run cost (USD) and wall-clock latency for \frameworkabbr{} and as relative multiples ($\times$) of \frameworkabbr{} for other methods. \frameworkabbr{} achieves the highest \passk{3} in all four settings while operating at substantially lower per-run cost. Batch2Code, a conversion-only variant of our framework is also efficient, but with substantially lower \passk{k}.
    }
    \label{tab:main_results_arxiv}
\end{table}

\label{sec:results:reliability}

\begin{figure}[t]
    \centering
    \includegraphics[width=0.9\textwidth]{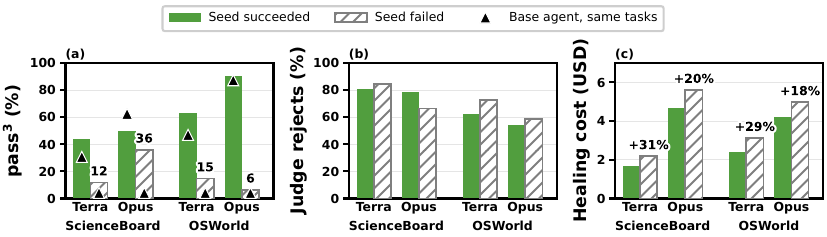}
    \caption{
        \frameworkabbr{} can recover reliable policies even from unsuccessful seed trajectories, but at higher cost.
        (a) \passk{3}, with base-agent performance on the same tasks.
        (b) Judge rejection rate during policy iteration.
        (c) One-time healing cost with the relative increase when initialization uses failed seeds.
    }
    \label{fig:results:initial_trajectory}
    \vspace{-10pt}
\end{figure}

\Cref{tab:main_results_arxiv} shows a clear reliability gap under repeated execution. Averaged across the four benchmark--model settings, \frameworkabbr{} drops only 3.4 points from \passk{1} to \passk{3}, compared with 12.4 points for the base agent, and achieves the highest \passk{3} in every setting, by 8.1 points on average over the next-best method. Its \passk{1} remains comparable to the base agent, so the flatter curve is not obtained by starting from a weaker policy. \frameworkabbr{} can also bootstrap from unsuccessful demonstrations. Across 356 tasks for which the base agent failed all three seed attempts, the final policy succeeds in \passk{3} on 62 tasks, including 32 of 89 on ScienceBoard with Opus (Figure~\ref{fig:results:initial_trajectory}). This recovery requires more refinement, with failed seeds incurring 18--31\% higher healing cost and 17--38\% longer healing time than successful seeds. We provide the full breakdown in Appendix~\ref{sec:appendix:failed_seeds}.

The comparison with Batch2Code helps distinguish reliable execution from simply making execution more deterministic. Batch2Code uses the same executable policy representation without iterative refinement, and its \passk{k} curve is similarly flat, but its \passk{3} remains 3.3–33.1 points below \frameworkabbr{}. Code therefore helps preserve a procedure across runs, while refinement makes it consistently correct. ASI, which instead retains an agent to select and execute learned programmatic skills, shows a decline with repeated execution closer to the base agent. Figure~\ref{fig:main_results_vis} illustrates the same trade-off on ScienceBoard with GPT-5.6 Terra. \frameworkabbr{} maintains its success rate as \(k\) increases while moving execution to a much lower-cost regime. Across all four settings, the resulting policies reduce per-run cost by 15-217x and run 3.4–5.1× faster than the corresponding base agents.

\newsavebox{\mainresultspanelbox}
\newcommand{\mainresultspanel}[2]{%
    \sbox{\mainresultspanelbox}{#2}%
    \leavevmode\usebox{\mainresultspanelbox}%
    \llap{\makebox[\wd\mainresultspanelbox][l]{\raisebox{\dimexpr\ht\mainresultspanelbox-\height\relax}{\footnotesize\bfseries(#1)}}}}
\begin{figure}[t]
    \centering
    \mainresultspanel{a}{\includegraphics[width=0.44\textwidth]{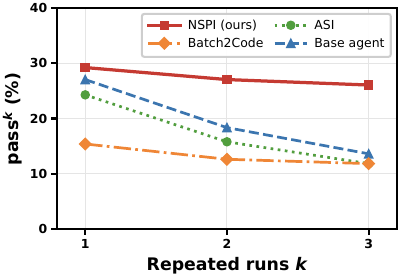}}\hfill
    \mainresultspanel{b}{\includegraphics[width=0.44\textwidth]{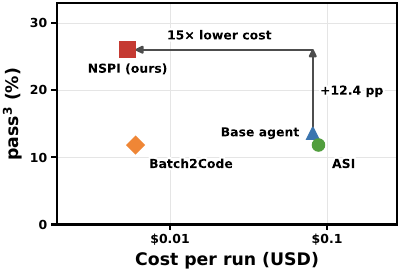}}
    \caption{
        Repeated-execution reliability and execution cost on ScienceBoard with GPT-5.6-Terra.
        (a) \frameworkabbr{} maintains high \passk{k} as k increases, while agent-based baselines degrade more sharply and the Batch2Code conversion-only baseline remains consistent but less accurate. (b) \frameworkabbr{} achieves highest \passk{3} with efficient per-run execution cost over baselines. Policy-construction cost is analyzed separately in \Cref{sec:results:cost_amortization}.
    }
    \label{fig:main_results_vis}
\end{figure}

\subsection{Policy generalization across workflow instances}
\label{sec:results:generalization}

\begin{figure}[t]
    \centering
    \small
    \setlength{\tabcolsep}{3.5pt}%
    \renewcommand{\arraystretch}{1.15}%
    \noindent
    \hbox to\textwidth{%
        \resizebox{0.50\textwidth}{!}{%
            \begin{tabular}{@{}p{0.72in}lccc@{}}
                \toprule
                & & \multicolumn{3}{c}{\textbf{Success rate (\%)}} \\
                \cmidrule(lr){3-5}
                \textbf{Setting} & \textbf{Method}
                    & \textbf{\passk{1}}
                    & \textbf{\passk{2}}
                    & \textbf{\passk{3}} \\
                \midrule

                \multirow{2}{0.72in}{\parbox{0.72in}{\raggedright Parameters (40)}}
                    & AutoRPA & 37.5 & 37.5 & 37.5 \\
                    & \textbf{\frameworkabbr{} (ours)} & \textbf{60.0} & \textbf{57.5} & \textbf{55.0} \\
                \addlinespace[0.4em]
                \multirow{2}{0.72in}{\parbox{0.72in}{\raggedright Initial states (35)}}
                    & AutoRPA & 55.2 & 52.4 & 51.4 \\
                    & \textbf{\frameworkabbr{} (ours)} & \textbf{61.0} & \textbf{60.0} & \textbf{60.0} \\
                \bottomrule
            \end{tabular}%
        }%
        \hfill
        \raisebox{-0.5\height}{%
            \includegraphics[width=0.45\textwidth]{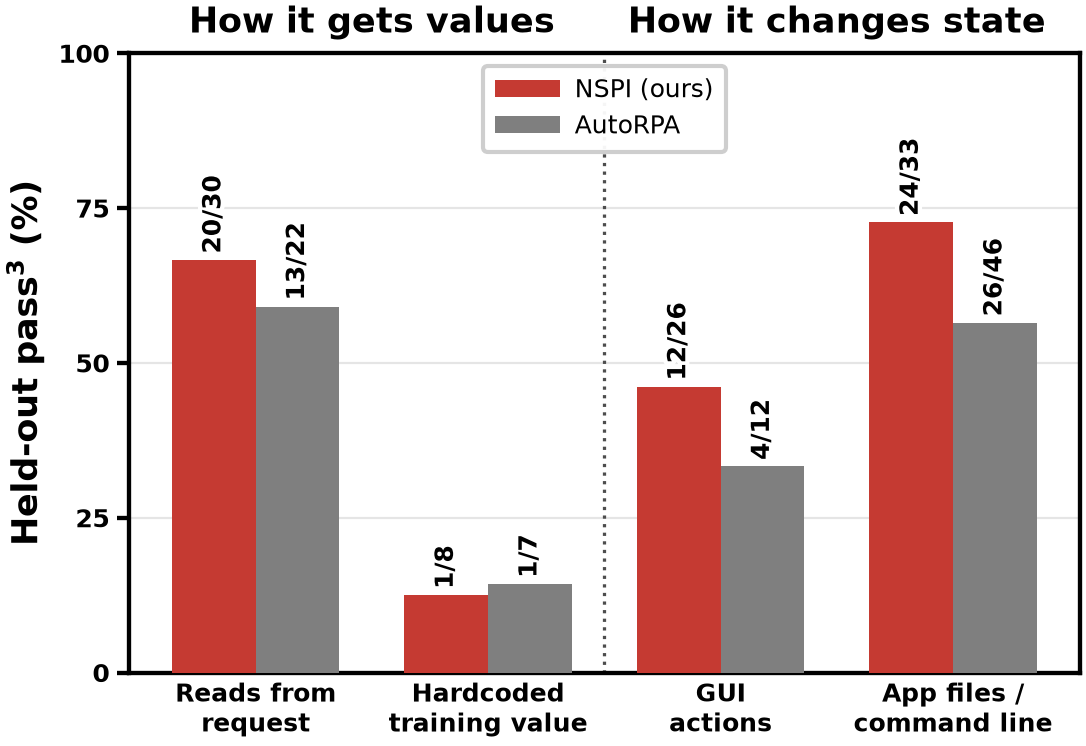}%

        }%
    }

    \vspace{0.35em}

    \begin{minipage}[t]{0.50\textwidth}
        \captionof{table}{
            Effectiveness of policy built across parameter and initial state variations to held-out instances of OSWorld workflows.
        }
        \label{tab:generalization}
    \end{minipage}\hfill
    \begin{minipage}[t]{0.45\textwidth}
        \captionof{figure}{
    Policy structures associated with successful reuse across workflow instances. Labels show the number of successful policies out of total applicable policies. 
        }
        \label{fig:generalization_curve}
    \end{minipage}
    \par\vspace{-15pt}
\end{figure}

We evaluate reuse across workflow instances on OSWorld with GPT-5.6 Terra, refining one policy on generated variants and evaluating the frozen policy on the held-out original task. As shown in Table~\ref{tab:generalization}, \frameworkabbr{} reaches 55.0\% \passk{3} on parameter variations and 60.0\% on initial-state variations, compared with 37.5\% and 51.4\% for AutoRPA. Both methods degrade a little from \passk{1} to \passk{3}, suggesting that the dominant challenge is whether a learned policy transfers to a workflow instance at all, rather than stochastic failure once it does. \frameworkabbr's advantage comes primarily from transferring successfully to more workflow instances.

Figure~\ref{fig:generalization_curve} examines which policy structures are associated with successful transfer. For parameterized workflows, policies that read changing values from the request transfer much more often than policies that retain training values, supporting the need to separate stable workflow structure from instance-specific inputs. For initial-state variation, policies that modify application files or use command-line interfaces transfer more often than GUI-only policies for both methods, consistent with being less dependent on the precise starting GUI state. AutoRPA relies on these script-based paths more heavily, yet \frameworkabbr{} performs better within both execution styles, so its advantage is not simply due to bypassing the GUI. Overall, successful reuse appears to depend on preserving the procedure while rebinding the parts that vary with the current request or environment. 

\subsection{Efficient repeated execution and cost amortization}
\label{sec:results:cost_amortization}
\vspace{-10pt}

We analyze the economics of neuro-symbolic policies using Eq. (\ref{eq:amortized}). Table \ref{tables/scienceboard_construction.tex} reports the resulting costs. Across different configurations, neuro-symbolic policy's per-execution cost ratio $c_\mu/c_\pi$ ranges from roughly 15$\times$ to roughly 217$\times$, and policies recover their construction cost between 3.1 and 26.8 executions. The break-even points in wall-clock time range between 8.6 and 16.9 executions.

The results also indicate that the benefit of NSPI scales with task complexity. Moving from ScienceBoard to OSWorld increases the agent's per-run dollar cost by four to six times, while construction cost increases by only 14\% to 52\%. The break-even point drops nearly 75\% in dollar value. The choice of a base model has a similar effect in dollar value. Claude Opus 5 incurs two to three times the construction cost of GPT-5.6 Terra, but its per-run agent cost is four to six times higher, so in dollars it breaks even roughly 2 to 2.4 times sooner. In wall-clock time, the two models are much closer. We note that these comparisons span only two benchmarks and two base models, so task complexity is partially confounded with other differences between benchmarks.
\vspace{-10pt}

\begin{table}[t]
\caption{
    Cost and wall-clock latency amortization of learning a reusable \frameworkabbr{} in the terms of \Cref{eq:amortized}. \(C_\pi\) is the one-time policy-construction cost, \(c_\mu\) the per-run cost of the base agent, \(c_\pi\) the per-run cost of executing the learned policy, and \(n^*\) is the break-even count. Values are means over tasks.
  } 
  \label{tables/scienceboard_construction.tex}
\begin{center}  
  \small
  \setlength{\tabcolsep}{3.5pt}
  \begin{tabular}{llrrrrrrrr}
  \toprule
  & & \multicolumn{4}{c}{\textbf{Cost (USD)}} & \multicolumn{4}{c}{\textbf{Wall-clock time}} \\
  \cmidrule(lr){3-6}\cmidrule(lr){7-10}
  \textbf{Benchmark} & \textbf{Base model} & $C_\pi$ & $c_\mu$ & $c_\pi$ & $n^\ast$ & $C_\pi$ (min) & $c_\mu$ (s) & $c_\pi$ (s) & $n^\ast$ \\
  \midrule
  \multirow{2}{*}{ScienceBoard} & GPT-5.6 Terra & 2.03 & 0.081 & 0.0054 & 26.8 & 40 & 183 & 40 & 16.9 \\
   & Claude Opus 5 & 5.68 & 0.52 & 0.0079 & 11.2 & 49 & 251 & 64 & 15.8 \\
  \midrule
  \multirow{2}{*}{OSWorld v1} & GPT-5.6 Terra & 3.09 & 0.48 & 0.0072 & 6.5 & 33 & 286 & 56 & 8.6 \\
   & Claude Opus 5 & 6.45 & 2.1 & 0.0097 & 3.1 & 43 & 353 & 102 & 10.3 \\
  \bottomrule
  \end{tabular}
    \vspace{-10pt}
\end{center}
\end{table}

\subsection{Extension to long workflows on OSWorld-V2}
\label{sec:results:longer-tasks}

\begin{rightwrapfig}[0.48\columnwidth]
    \small
    \setlength{\tabcolsep}{3.5pt}%
    \renewcommand{\arraystretch}{1.15}%
    \begin{tabular}{@{}lccc@{}}
        \toprule
        \textbf{Method}
            & \textbf{\passk{1}}
            & \textbf{\passk{2}}
            & \textbf{\passk{3}} \\
        \midrule
        \multicolumn{4}{@{}l}{\textit{Partial credit}} \\
        Base agent    & \textbf{72.79} & \textbf{69.10} & \textbf{63.09} \\
        NSPI & 45.37          & 44.83          & 44.60          \\
        \midrule
        \multicolumn{4}{@{}l}{\textit{Strict success}} \\
        Base agent    & \textbf{28.04} & \textbf{25.23} & \textbf{22.43} \\
        NSPI & 17.45          & 17.13          & 16.82          \\
        \bottomrule
    \end{tabular}

    \captionof{table}{
        \passk{k} performance on OSWorld-V2. 
    }
    \label{tab:osworld_v2}
\end{rightwrapfig}
\noindent
We stress-tested \frameworkabbr{} on OSWorld-V2~\citep{yuan2026osworld2}, where tasks take skilled humans more than an hour on average and cost frontier-model agents up to \$76 and 50 minutes per task. \Cref{tab:osworld_v2} shows that the learned policies are more consistent than the base agent but less successful. On strict success, \frameworkabbr{} keeps 96\% of its \passk{1} at $k{=}3$, whereas the base agent keeps only 80\%. However, \frameworkabbr{} trails the base agent at $k{=}3$ by 5.6 points in strict success and by 18.5 points in partial credit. The gap in performance arises from a loss of information in converting a long agent trajectory to initial policy: agent trajectories averaged 158 steps and scripts averaged 970 lines, leading to mistakes when the script recomputes every value the agent read from the screen. Implementation details are available in \Cref{sec:appendix:osworld-v2}.

\subsection{Pre-action verification under environment variation}
\label{sec:results:precondition}

We evaluate the pre-action verifier (\Cref{sec:method:robustness:verifier}) with the injected intent comments and rubric additions of Appendix~\ref{sec:appendix:precondition} at $\theta = 0.78$. The \emph{wrong-block rate} (WBR) is the fraction of rollouts of correct policies in which the verifier incorrectly blocks at least one primitive. The \emph{wrong-pass rate} (WPR) is the fraction of primitives that should be blocked but that the verifier passes. We simulate an unexpected environment change by running the correct policies of original OSWorld tasks but starting from the perturbed initial states of the generalization experiment (\Cref{sec:results:generalization}), which yields 24 primitives that should be blocked. \Cref{fig:precondition_verifier} shows that Qwen3.6-35B-A3B wrongly blocks at most 8.4\% of correct rollouts in every setting, while Qwen3.8-27B wrongly blocks 9.4--25.0\%. The trade-off reverses on the perturbed initial states: Qwen3.6-35B-A3B has WPR 70.8\% (17 of 24 primitives passed), while Qwen3.8-27B has WPR 25.0\% (6 of 24). Median verifier latency is $1.5\text{s} \pm 0.5$ (p95 $2.3\text{s}$) for Qwen3.6-35B-A3B and $1.6\text{s} \pm 1.2$ (p95 $3.7\text{s}$) for Qwen3.8-27B.

\subsection{Judge Analysis}
\label{sec:results:ablations}

\label{sec:results:judge}

\Cref{tab:results:judge_alignment} compares our judges with the benchmark evaluator on each task's final policy, where a policy is reliable only if it succeeds in all three runs. The judges reject 69.3--94.4\% of unreliable policies but accept only 11.4--40.1\% of reliable ones. They are therefore conservative, which increases construction cost without hurting reliability. A better-calibrated judge could lower this cost and potentially improve performance, which we leave to future work.

We also annotate 19 OSWorld policies rejected by $J_\text{task}$ and compare both judges with human labels (\Cref{sec:appendix:judge_analysis}). \Cref{tab:results:judge_analysis} shows $J_\text{task}$ agrees closely with human annotators, and so does $J_\text{loc}$ on divergence localization when the evidence is visible on screen. We discuss four limitations of the judges in the same appendix.

\begin{table}
    \small
    \setlength{\tabcolsep}{3.5pt}%
    \renewcommand{\arraystretch}{1.15}%
    \begin{minipage}[c]{0.48\textwidth}
        \centering
        \begin{tabular}{@{}llrrr@{}}
            \toprule
            \textbf{Benchmark} & \textbf{Models} & \textbf{Prec.} & \textbf{Rec.} & \textbf{Spec.} \\
            \midrule
            \multirow{2}{*}{ScienceBoard} & Terra & 41.7 & 11.4 & 94.4 \\
             & Opus 5 & 69.6 & 22.2 & 92.7 \\
            \midrule
            \multirow{2}{*}{OSWorld} & Terra & 76.8 & 40.1 & 87.6 \\
             & Opus 5 & 76.7 & 35.3 & 69.3 \\
            \bottomrule
        \end{tabular}
    \end{minipage}\hfill
    \begin{minipage}[c]{0.48\textwidth}
        \centering
        \begin{tabular}{@{}lr@{}}
            \toprule
            \textbf{Judge output} & \textbf{Agreement} \\
            \midrule
            $J_\text{task}$ verdict & 84\% (16/19) \\
            $t^\ast$, visible failures & 90\% (9/10) \\
            $t^\ast$, invisible failures & 0\% (0/6) \\
            \bottomrule
        \end{tabular}
    \end{minipage}

    \vspace{0.5em}
    \begin{minipage}[t]{0.48\textwidth}
        \captionof{table}{
        Agreement between \frameworkabbr{} judges and the benchmark evaluator on final policies.
        }
        \label{tab:results:judge_alignment}
    \end{minipage}\hfill
    \begin{minipage}[t]{0.48\textwidth}
        \captionof{table}{
            Agreement of \frameworkabbr{} judges with human annotations on 19 OSWorld tasks.
        }
        \label{tab:results:judge_analysis}
    \end{minipage}
\end{table}

\section{Related work}
\label{sec:related}

Prior methods on learning skills from experience enable an agent to accumulate a library of reusable APIs, represented either in natural language \citep{zheng2025skillweaver,wang2025agent} or as short programs \citep{wang2023voyager,meng2025growing,wang2025inducing,ning2026code}. We take this approach to the limit by aiming for a single robust code policy that solves each equivalence set of parametric tasks in full. While the neural agent is the primary driver that invokes skills in prior work, our neuro-symbolic policy is primary while the agent is an expensive fallback.

Programs with perceptual inputs have long served as policies, from programmatic reinforcement learning~\citep{verma2018programmatically} to LLM-generated robot code~\citep{liang2023code,santos2026alrm}. The LLM-generated policies were produced from human natural language instructions and few-shot examples, instead of from a strong agent exploration as in our work. They were also static, whereas our neuro-symbolic iterative refinement method allows the policy to improve over time based on execution feedback and continual agent exploration.

The most similar prior work is AutoRPA \citep{chen2026autorpa}, which converts ReAct-style agent exploration into replayable RPA scripts with iterative refinement, demonstrating substantial cost reduction while maintaining baseline pass@1 success rate. Our work differs in that our method explicitly optimizes \passk{k} success rate and realistic intra-task generalization to different task parameters and initial states, along with a precondition check for practical deployment.

\section{Conclusion}
\label{sec:conclusion}
We introduced neuro-symbolic computer use as a way to turn execution experience into reusable policies for recurring workflows, preserving stable procedure in code while leaving state-dependent decisions to neural primitives. Neuro-Symbolic Policy Iteration learns these policies from agent trajectories through execution, diagnosis, exploration, and revision, and further refines them across task-parameter and initial-state variations. Across OSWorld and ScienceBoard, the resulting policies improve reliability under repeated execution while substantially reducing per-run cost and latency, and they transfer more successfully than AutoRPA to held-out instances of the same workflow. Results on longer-horizon tasks and our judge and verifier analyses also expose remaining bottlenecks in trajectory-to-policy conversion and policy evaluation, suggesting that better acquisition and verification could further extend this approach to more complex recurring workflows.

\bibliography{iclr2027_conference}

\appendix

\section{Method details}
\label{sec:appendix:method}

\subsection{Primitive action space}
\label{sec:appendix:action_space}

We instantiate policies in Simulang~\citep{simulang}, a scripting language for desktop and web automation in the style of PyAutoGUI~\citep{pyautogui} and Playwright~\citep{playwright} that also provides the neural primitives above. \Cref{tab:action_space} lists the primitives that the converter and the healer can use on ScienceBoard. The healer can combine them with Python control flow, e.g., a loop that retries a click or a branch on \texttt{state\_satisfies}.

\begin{table}[t]
    \caption{
        Primitive action space of policies on ScienceBoard.
        Each primitive is a method of an \texttt{agent} object that the policy calls, e.g., \texttt{agent.click("File menu")}.
        Neural primitives call a model on the current observation $o_t$: the grounded actions use UI-Venus-1.5-30B-A3B \citep{team2026ui} with a consensus over three parallel calls, and the condition check uses Gemini~3 Flash.
        A check mark under \emph{Mut.} marks a primitive that changes the state of the computer; these count toward the per-task step budget and are gated by the pre-action verifier.
    }
    \label{tab:action_space}
    \centering
    \small
    \renewcommand{\arraystretch}{1.1}
    \newcommand{\primgroup}[1]{\multicolumn{3}{@{}l}{\textit{#1}} \\}
    \begin{tabular}{@{}p{0.37\textwidth} c p{0.52\textwidth}@{}}
        \toprule
        \textbf{Primitive} & \textbf{Mut.} & \textbf{Behavior} \\
        \midrule
        \primgroup{Neural: grounded GUI action $G(\ell, o_t)$}
        \texttt{click(}$\ell$\texttt{, clicks, button, hold\_keys)} & \checkmark & Grounds the element description $\ell$ to a screen coordinate and clicks it; \texttt{clicks} sets double or triple clicks and \texttt{hold\_keys} holds modifiers. \\
        \texttt{type(}$\ell$\texttt{, text, enter, overwrite)} & \checkmark & Clicks the element grounded from $\ell$ if given, clears it if \texttt{overwrite}, types \texttt{text}, and presses Enter if \texttt{enter}. \\
        \texttt{scroll(}$\ell$\texttt{, amount)} & \checkmark & Moves the pointer to the element grounded from $\ell$ and scrolls by \texttt{amount}. \\
        \texttt{drag\_and\_drop(}$\ell_\text{start}$\texttt{, }$\ell_\text{end}$\texttt{, hold\_keys)} & \checkmark & Grounds both descriptions and drags from the first element to the second. \\
        \midrule
        \primgroup{Neural: condition check $Q(\phi, o_t)$}
        \texttt{state\_satisfies(}$\phi$\texttt{)} $\rightarrow$ \texttt{bool} & & Returns whether the current screenshot satisfies the natural-language condition $\phi$; the plan can branch on the result. \\

        \midrule
        \primgroup{Symbolic: keyboard and applications}
        \texttt{hotkey(keys)} & \checkmark & Presses a key combination, e.g., \texttt{["ctrl", "s"]}. \\
        \texttt{hold\_and\_press(hold, press)} & \checkmark & Holds the keys in \texttt{hold} while pressing the keys in \texttt{press}. \\
        \texttt{open(path)} & \checkmark & Opens a file with its default application, or launches an application by name. \\
        \midrule
        \primgroup{Symbolic: programmatic}
        \texttt{\_exec\_python(code)} $\rightarrow$ \texttt{str} & \checkmark & Runs Python on the computer and returns its standard output. \\
        \texttt{\_exec\_bash(cmd)} $\rightarrow$ \texttt{str} & \checkmark & Runs a shell command on the computer and returns its standard output. \\
        \midrule
        \primgroup{Symbolic: control}
        \texttt{wait(seconds)} & & Waits for up to 30 seconds, e.g., for a dialog to open. \\
        \texttt{done()} & & Ends the plan and reports the task as complete. \\
        \texttt{answer(x)} & & Ends the plan and submits \texttt{x}, which the evaluator matches exactly, for question-answering tasks. \\
        \texttt{fail()} & & Ends the plan and reports the task as infeasible. \\
        \bottomrule
    \end{tabular}
\end{table}

\subsection{Initialization}
\label{sec:appendix:method:init}

The base agent $\mu$ runs on $g$ in exploration mode, in which actions may be arbitrary code, and we record its trajectory $\tau_\mu$. The converter $\Phi_\text{conv}$ first extracts the discrete actions of $\tau_\mu$ into a single contiguous script (Section~\ref{sec:experiment_setup}). It then replaces pixel coordinates, which were recorded on a different machine and are unreliable, with natural-language descriptions of the target elements that the runtime grounds at execution time, and it replaces values specific to $g$ with task parameters. On ScienceBoard, this second step is one LLM call that receives the task, the extracted script, and a screenshot and accessibility tree of the initial state. On OSWorld, the base agent trajectory was based on Agent-S3~\citep{agents3} which already encodes the trajectory in semantic-level rather than pixel-level, then calls an LLM for the final conversion. Conditioning $\Phi_\text{conv}$ on $g$ lets it drop incidental actions in $\tau_\mu$, such as backtracking, exploratory clicks, and redundant verifications, so that they are not transcribed into the code policy that will be refined.

\subsection{Iteration}
\label{sec:appendix:method:loop}

The rollouts within an iteration run in a random order, and the iteration stops at the first rollout that fails the judge, so the agent continuation runs at most once per iteration. Every code rollout and every agent continuation is capped at the task's step budget, i.e., the maximum number of state-mutating actions, which matches the budget of the baseline agent; the localizer and the healer are told the budget so that their suggestions fit within it. The agent continuation can use GUI actions and run Python or shell commands on the machine, and it receives the output of those commands in later turns. The continuation is not judged and does not change the verdict of the code rollout; it serves only as experience for the healer. A rollout that raises a code execution error (a syntax error, an undefined name, a failed import, etc.) produces no trajectory for the judges to grade, so we skip the judges and the continuation and give the traceback to the healer. The healer receives at most five screenshots of the failed rollout, sampled evenly over the trajectory, and it rewrites the whole code policy. The converter and the healer are instructed to prefer programmatic actions (shell commands, Python, and application command lines) over long sequences of GUI actions, which replaces neural primitives with symbolic ones and lowers the cost per execution $c_\pi$, although our method does not optimize cost directly.

\subsection{Dual judge}
\label{sec:appendix:judges}

\begin{algorithm}[t]
\caption{Rollout diagnosis with the dual judge (policy evaluation)}
\label{alg:diagnose}
\begin{algorithmic}[1]
\Require rollout $\tau = (o_0, a_0, \ldots, o_{|\tau|})$; task $g$; policy $\pi$; task-completion judge $J_\text{task}$; observer $J_\text{obs}$; localizer $J_\text{loc}$; applicability gate $\textsc{Applicable}$; aggregator $\textsc{Aggregate}$
\Procedure{Diagnose}{$\tau, g, \pi$}
\State $(v_\text{task}, e_\text{task}) \gets (\text{skipped}, \emptyset)$
\If{$\textsc{Applicable}(g)$} \Comment{is success checkable from persistent state?}
\State $(v_\text{task}, e_\text{task}) \gets J_\text{task}(g)$ \Comment{read-only probes of the end state, in parallel with the step-level judge}
\EndIf
\For{$t = 0, 1, \ldots, |\tau| - 1$} \textbf{in parallel} \Comment{blind to $g$, $\pi$, and step intent}
\State $\delta_t \gets J_\text{obs}(o_t, o_{t+1})$
\EndFor
\State $(v_\text{loc}, t^\ast, m) \gets J_\text{loc}(g, \pi, \delta_{0:|\tau|-1})$ \Comment{verdict, first divergent step, failure mode}
\State $v \gets \text{success}$ \textbf{if} $v_\text{loc} = \text{success}$ \textbf{and} $v_\text{task} \in \lbrace \text{success}, \text{skipped} \rbrace$ \textbf{else} failure \Comment{abstain counts as failure}
\State \Return $\big(v, \textsc{Aggregate}(v_\text{task}, e_\text{task}, v_\text{loc}, t^\ast, m)\big)$ \Comment{verdict and diagnostic $d$}
\EndProcedure
\end{algorithmic}
\end{algorithm}

\textbf{Motivation for two judges.} Evaluating a computer-use rollout is harder than reading an exit code, for two reasons. First, an agent's own report of what it did is often optimistic, e.g., it may report that it clicked a button while the screen shows that the click missed, or that it entered a value into a spreadsheet cell while it edited the wrong cell, so an evaluator that trusts the actor's narrative inherits its errors. Second, benchmark evaluation scripts such as those of OSWorld check the end state of the environment (the contents of a saved spreadsheet, an exported image, the browser's bookmarks, etc.), which an evaluator that only watches the trajectory cannot see. For example, a LibreOffice Calc plan that types the correct formula but closes the window before the file is saved, a GIMP plan whose edit is never exported to the target file, and a Chrome plan whose bookmark fails to persist all look correct frame by frame, yet all fail the evaluation script. The task completion judge addresses the second problem, and the step-level judge addresses the first while also localizing the error for the healer. We ablate the two judges in Section~\ref{sec:results:ablations}.

\textbf{Task completion judge.} $J_\text{task}$ is a VLM agent with read-only probes of the live machine: shell commands, file reads, directory listings, SQL queries over application databases, and screenshots. Mutating shell commands are rejected by a deny-list, and databases are opened read-only. $J_\text{task}$ receives only the task description (and the submitted answer for question-answering tasks), and never another judge's verdict or the benchmark's verdict. It decomposes the goal into verifiable sub-conditions, probes the machine until it can settle each one (preferring filesystem and database probes over screenshots), and returns a verdict with supporting evidence within a budget of 30 tool calls and 300 seconds. It abstains (\texttt{cannot\_verify}) when the state needed to check success is transient or depends on context that is not persisted, e.g., for ``move the file back where it used to be'', the original path no longer exists. $J_\text{task}$ is not a ground-truth oracle, because its verdict is only as good as the probes it chooses to run; this is why an abstention falls back to the step-level judge.

\textbf{Applicability gate.} One VLM call on the task instruction, cached per task, decides whether success can be checked from persistent machine state. If not, $J_\text{task}$ is skipped and the verdict comes from the step-level judge alone, which spends no probe budget on a question $J_\text{task}$ could not answer.

\textbf{Observers.} For GUI primitives, the evidence for transition $t$ is the pair of screenshots before and after the step, together with the difference between accessibility trees when available. For primitives that leave no visual trace, e.g., API calls, background shell commands, or functions executed inside the code runtime, the evidence is the return value, exit code, and captured output, and no screen change is expected. Each observer returns an objective description of what the step produced and an anomaly grade. Observers do not see the task, the code, or the stated intent of the step, because an observer told that the agent intended to save a file can no longer distinguish ``the agent believes it saved'' from ``the file was saved'', which is the distinction the evaluation script checks. The $|\tau|$ observer calls are independent and run in parallel.

\textbf{Localizer.} The localizer is one VLM call that receives the task, the executed code, and the ordered observer outputs, but not the actor's narrative. It enumerates the sub-goals of the task and checks each against the observer evidence. It also screens for composition failures, i.e., sequences of individually plausible steps that together are incoherent or destructive, such as an action that is immediately undone, a target value that is overwritten, or a save followed by closing the application before the write completed. It returns the verdict, the first divergent step $t^\ast$, and a name for the failure mode, which the healer uses when rewriting the code policy.

\textbf{Diagnostic.} An LLM composes the outputs of both judges into the diagnostic $d$. $d$ lists each judge's observations (the probe results of $J_\text{task}$ and the observer descriptions) separately from its interpretations (verdicts, failure hypotheses, and suggested focus), and it tells the healer that observations are facts about the machine while verdicts can be wrong.

\subsection{Pre-action verifier}
\label{sec:appendix:verifier}

\begin{figure}[t]
\centering
\includegraphics[width=1.0\linewidth]{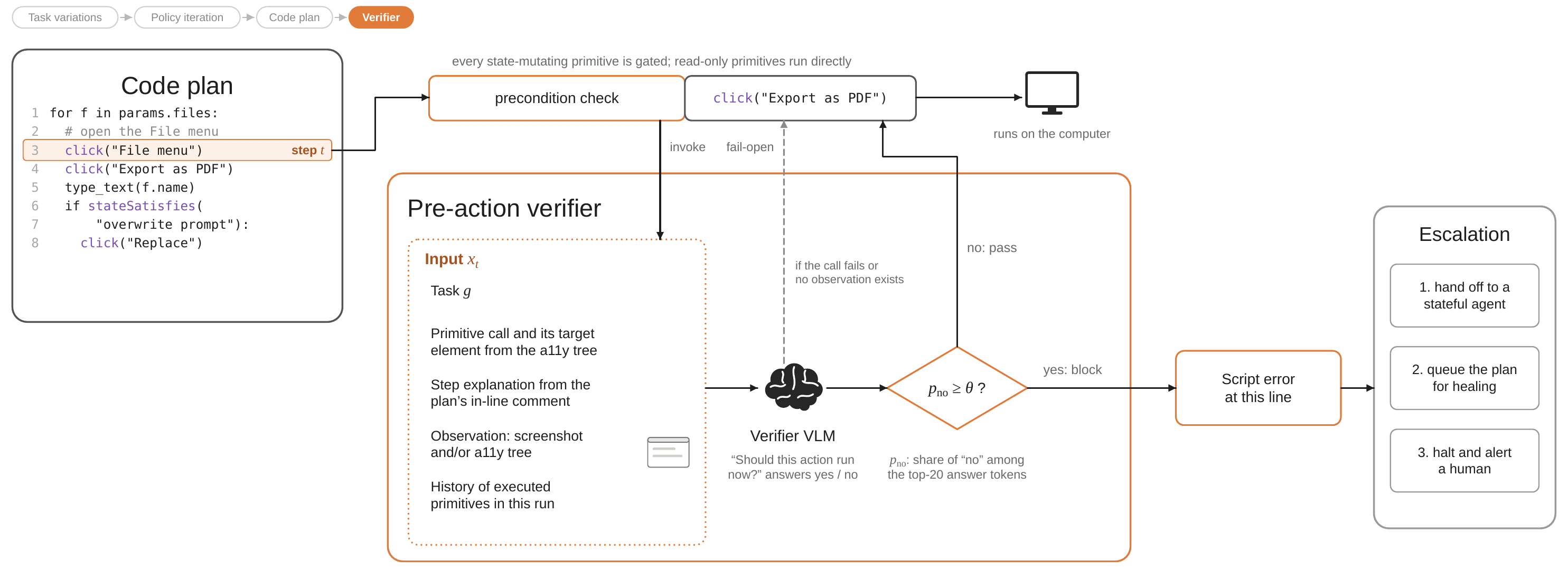}
\caption{
A learned binary-output pre-action verifier $V$ runs before every state-mutating primitive during policy deployment. Reusing the plan's most recent stashed screenshot and accessibility tree, $V$ emits top-20 answer-token logprobs; the yes/no log-mass is renormalized to $p_\text{no}$ and the action is blocked iff $p_\text{no} \geq \theta$. A block raises a script-level error at the offending line, triggering the fallback hierarchy (agent hand-off, code repair, halt).
}
\label{fig:verifier}
\end{figure}

\textbf{Verifier and safety guardrails.} A safety guardrail confirms that an irreversible action is intended, e.g., a final \texttt{click(\{concept: "Pay now"\})}, and is invoked at a small number of known points. The pre-action verifier gates every state-mutating primitive, including reversible ones, against the implicit assumptions under which the line of code was written. A click on an email field is reversible, but if a cookie-consent banner is absorbing pointer events, the click is lost and every later step operates on an unchanged page. We therefore do not skip the check for reversible actions, since a downstream component that could undo a wrong execution may not exist.

\textbf{Verifier and in-code state checks.} A code policy may also contain its own state checks, e.g., a \texttt{stateSatisfies(\{condition\})} line inserted where the planner expects a particular state. Three properties separate these from the verifier. First, in-code checks run only where the planner placed them, whereas the verifier runs before every gated primitive. Second, an in-code check carries whatever condition string the planner wrote, which differs across plans and cannot be improved without rewriting the plan, whereas the verifier uses one fixed prompt that we version separately from any code policy. Third, the verifier exposes the threshold $\theta$, whereas an in-code check returns a hard yes or no. We treat in-code checks as a planner affordance for control flow, and they do not replace the verifier.

\textbf{Pre-action and post-action checks.} Any error that a post-action check on $a_t$ could detect, i.e., a primitive that returned successfully without its intended effect, is also detectable by the pre-action check on $a_{t+1}$, because the precondition of $a_{t+1}$ was written against the intended post-state of $a_t$. A pre-action check therefore covers post-action checking up to a lag of one step, uses the same observation as the next action, and leads directly to the fallback, since the response to a failed check is to not execute the next primitive. The final primitive of a plan has no successor to catch its failed effect; the runtime reserves a post-action slot for an outcome check there, and end-of-rollout verification is provided by the task completion judge.

\textbf{Gated primitives.} We gate primitives that write to the state of the computer: pointer and keyboard input, form mutation, navigation, filesystem and clipboard writes, integration writes such as spreadsheet cell updates, and in-page script execution, whether called directly or as a method on an application or page handle. In the ScienceBoard and OSWorld runtime these are \texttt{click}, \texttt{double\_click}, \texttt{type}, \texttt{scroll}, \texttt{hotkey}, \texttt{hold\_and\_press}, \texttt{open}, \texttt{drag\_and\_drop}, \texttt{set\_cell\_values}, and Python and shell execution. We gate reversible write primitives too, because an undo path may not exist (e.g., a click that triggers an analytics event), and because a skilled human operator would not execute a step against an obviously wrong state regardless of whether the step is reversible. We do not gate read-only primitives (screenshots, accessibility snapshots, reading page content or the clipboard), because they cannot corrupt external state and their outputs are the observations that the verifier consumes. We also do not gate primitives confined to the runtime (variable assignment, control flow, in-memory list operations), whose outcome is determined by the code and cannot disagree with the external world, or window and tab lifecycle operations, which typically run before any observation exists. When membership is ambiguous we gate the primitive, because a spurious check costs one verifier call that usually passes at a low wrong-block operating point. Writes issued through headless APIs, e.g., sending mail through a mail-service API, present no on-screen state to verify against and need a separate safeguard.

\textbf{Inputs.} Before executing a gated primitive at step $t$, the runtime builds the input
\begin{align}
\label{eq:verifier_input}
x_t = (g, c_t, e_t, o_t, h_t) ,
\end{align}
where $c_t$ is the code of the primitive call; $e_t$ is a natural-language explanation of the step taken from the in-line comments of the code policy; $o_t$ is the current observation, as a screenshot, an accessibility tree, or both; and $h_t$ lists the primitives already executed in this run. The verifier $V_\theta$ maps $x_t$ to a decision in $\lbrace \text{pass}, \text{block} \rbrace$. The explanation $e_t$ is the trailing comment on the statement that contains the call, otherwise the comment block directly above it, otherwise the first line of the enclosing function's docstring, and otherwise a literal restatement of the call. A verifier note supplies instance-specific knowledge that the verifier cannot infer, e.g., a machine-specific path or the operating system; the note is presented as a fact about the environment and not as approval, so the rubric still applies. The history $h_t$ is a compact list of the steps completed so far in the run, which lets the verifier distinguish a repeated action from its first occurrence without a full transcript.

\textbf{Log-probability aggregation.} We request the 20 most likely next tokens with their log-probabilities at the answer position, skipping any reasoning block that precedes the answer. Let $\ell_t(w)$ be the log-probability of token $w$ at the answer position, restricted to the 20 most likely tokens, and let $\Wcal_\text{yes}$ and $\Wcal_\text{no}$ be the surface forms of each answer (e.g., \texttt{yes}, \texttt{Yes}, \texttt{Y}). The probability of blocking is the renormalized mass of the no forms, and the verifier blocks when it reaches a threshold $\theta$ that is set by a calibration set (see Calibration below):
\begin{align}
\label{eq:pno}
p_\text{no}(x_t) = \frac{\sum_{w \in \Wcal_\text{no}} \exp \ell_t(w)}{\sum_{w \in \Wcal_\text{yes} \cup \Wcal_\text{no}} \exp \ell_t(w)} ,
\qquad
V_\theta(x_t) =
\begin{cases}
\text{block} & \text{if } p_\text{no}(x_t) \geq \theta , \\
\text{pass} & \text{otherwise} .
\end{cases}
\end{align}
The yes forms are \texttt{yes}, \texttt{Yes}, \texttt{YES}, \texttt{y}, \texttt{Y} and their variants with a leading space, and the no forms are defined likewise. We compute the mass of each class with a log-sum-exp over its forms and renormalize over the two classes to obtain $p_\text{no}$ (\Cref{eq:pno}). If neither class appears among the 20 tokens, we use the greedy answer, and if the completion contains neither answer, the verifier fails open.

\textbf{Calibration.}
\label{sec:method:verifier:calibration}
A user who automates a repetitive task expects a working code policy to run without interruption, so a wrong block (stopping a correct primitive) costs more than a wrong pass (running an incorrect primitive), as long as safety guardrails are in place. We therefore set $\theta$ in a calibration phase before any block is enforced. We run the verifier in shadow mode, which computes and logs $p_\text{no}$ for every gated primitive without enforcing blocks, on a calibration set of rollouts of final policies, and we label each rollout with its task outcome. We treat every primitive of a successful rollout as correct, so any block within a successful rollout is a wrong block. Let $\mathrm{WBR}(\theta)$ be the fraction of successful rollouts in which at least one primitive has $p_\text{no} \geq \theta$, and $\mathrm{DR}(\theta)$ the fraction of failed rollouts in which at least one primitive has $p_\text{no} \geq \theta$. Both rates are non-increasing in $\theta$, so for a wrong-block budget $\epsilon$ chosen by the deployment, the threshold with the highest detection rate within the budget is the smallest feasible threshold on a grid $\Theta$:
\begin{align}
\label{eq:theta}
\theta^\ast = \min \lbrace \theta \in \Theta : \mathrm{WBR}(\theta) \leq \epsilon \rbrace .
\end{align}

\section{Experiment details}
\label{sec:appendix:experiment-details}

\subsection{Main experiment configuration}
\label{sec:appendix:repeated-execution}

\textbf{Methods.} Both backbones, GPT-5.6 Terra and Claude Opus~5, are called through OpenRouter, and every method on a benchmark uses the same backbone in every LLM role except the judges, the \texttt{state\_satisfies} checker, and the continuation summarizer (\Cref{tab:appendix:main-config}). The \emph{base agent} is Agent-S3~\citep{agents3} on OSWorld, with UI-Venus-1.5-30B~\citep{team2026ui} as its grounding model, and ScienceBoard's own screenshot agent on ScienceBoard. \emph{\frameworkabbr{}} starts from one base-agent trajectory per task (\Cref{sec:appendix:method:init}) and runs policy iteration with $N{=}5$ iterations of up to $K{=}5$ rollouts. \emph{Batch2Code} converts three base-agent trajectories per task into a policy with one call of the same converter and no refinement. \emph{ASI}~\citep{wang2025inducing} builds its skill library in one online pass over the tasks on top of the same base agent; its later trials replay the agent with the library frozen after that task. Code policies of \frameworkabbr{} and Batch2Code share the grounding model and run under the same step limits.

\textbf{Scoring.} Each method gets three fresh trials per task, from which we estimate \passk{k} for $k\in\{1,2,3\}$ over the full task set. Cost is the provider-billed model cost plus the grounding model at \$3.25 per GPU-hour of call time. One-time construction cost covers conversion and policy iteration (\frameworkabbr{}), conversion (Batch2Code), or the online library-building pass (ASI); per-run cost covers one final trial.

\begin{table}[t]
\centering
\small
\setlength{\tabcolsep}{4pt}
\begin{tabular}{@{}lll@{}}
\toprule
\textbf{Setting} & \textbf{ScienceBoard} & \textbf{OSWorld} \\
\midrule
Base agent: output tokens, step limit & 8192; task step limit & 4096; 100 steps \\
Code rollout step limit & task limit $\times$5 (Terra), $\times$16 (Opus) & 100 primitive calls \\
Converter / healer / continuation tokens, Terra & 2000 / 4096 / 1024 & 4096 / 8192 / 1024 \\
Converter / healer / continuation tokens, Opus~5 & 4096 / 16384 / 2048 & 4096 / 16384 / 2048 \\
Continuation turns & 15 & 15 \\
\midrule
Observer / localizer ($J_\text{loc}$) & \multicolumn{2}{l}{Gemini 2.5 Flash / Gemini 2.5 Pro} \\
Task-completion judge $J_\text{task}$ & \multicolumn{2}{l}{Claude Sonnet 4.6 (30 tool calls, 300\,s)} \\
Applicability gate / judge combination & \multicolumn{2}{l}{Gemini 2.5 Pro / Gemini 2.5 Flash} \\
\texttt{state\_satisfies}, continuation summarizer & \multicolumn{2}{l}{Gemini 3 Flash} \\
\bottomrule
\end{tabular}
\caption{Main-experiment configuration. All judge roles use temperature 0.}
\label{tab:appendix:main-config}
\end{table}

\subsection{Workflow Variation}
\label{sec:appendix:task-variants-baseline}

\subsubsection{Task variant generation}
\label{sec:appendix:variant_generation}

\begin{figure}[t]
\centering
\includegraphics[width=\linewidth]{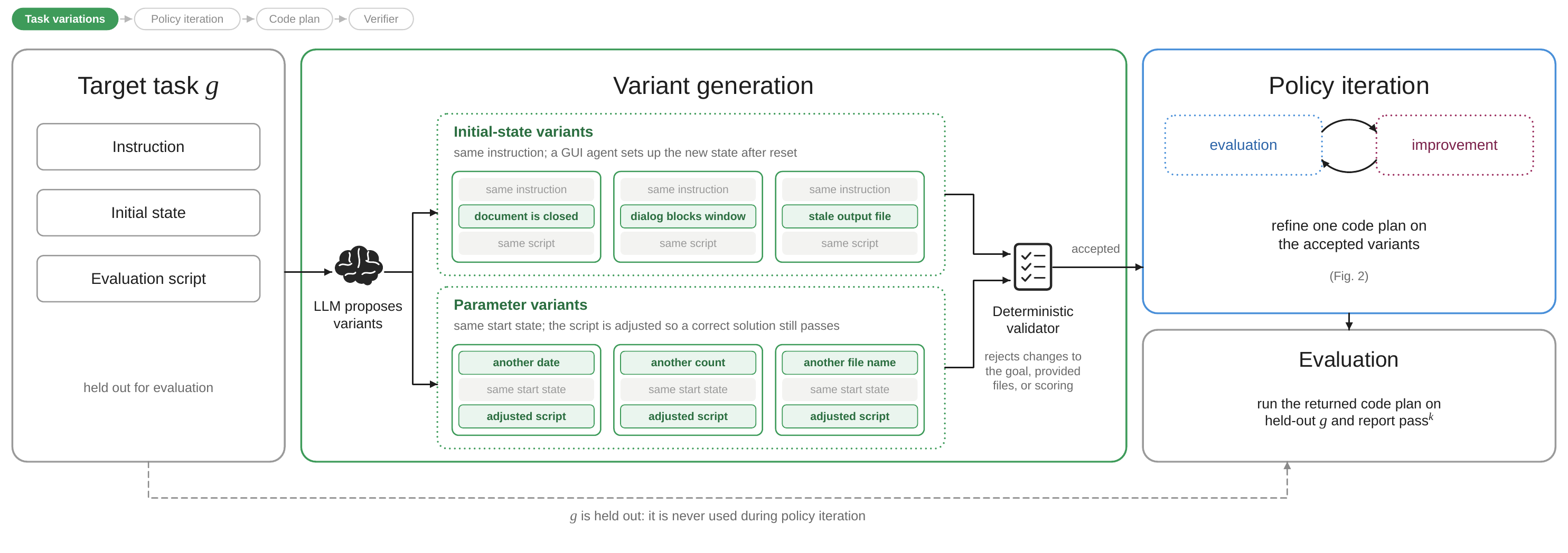}
\caption{
Given an original target task and environment configuration (initial state and evaluation script), we call an LLM to generate synthetic task variants that differ in the initial state and task parameters
}
\label{fig:variant_generation}
\end{figure}

We generate the variants of each task with one LLM call followed by a deterministic validator (\Cref{fig:variant_generation}). The LLM receives the full task specification, i.e., the instruction, the environment setup, and the evaluation script, together with notes that we extract from the evaluation script, e.g., whether it compares the output against a pre-rendered reference file or against inline constants. It first decides for each category whether the task admits variants at all, and then proposes up to three variants per admissible category. An initial-state variant keeps the instruction and adds a setup instruction that a general GUI agent executes before the policy runs, e.g., ``Close the spreadsheet window entirely, leaving only the file manager open.'' The setup instruction may only change transient application state (open windows, selected tabs, dialogs, etc.) or pre-fill the region that the task writes to, so the files provided by the task setup and the original solution stay unchanged. A parameter variant rewrites the instruction with new argument values and rebinds the constants in the evaluation script to match. The validator rejects a parameter variant if its instruction is unchanged, if the rebound evaluation script uses a different scoring function, or if the script still compares a changed output against the original reference file, because a correct solution could never match that file. It marks variants that pass but carry a warning, e.g., an edited evaluation script on an initial-state variant, for human review. Because the LLM may decline a category, parameter variants are much rarer than initial-state variants: on OSWorld, 332 of 369 tasks pass a pre-filter that removes infeasible tasks, all 332 admit initial-state variants, and only 93 admit parameter variants, mainly because 161 of the other 239 evaluation scripts compare against reference files that we cannot regenerate. We manually review the generated variants and remove infeasible cases. We retain only families whose original task was solved at least once by the Terra or Opus~5 base agent, so failures measure transfer rather than task infeasibility. The resulting benchmark contains 40 parameter families and 35 initial-state families across ten OSWorld domains, generally with three variants per family.

\textbf{Construction protocol.} Both methods use GPT-5.6 Terra for all construction roles and the same grounding model with UI-Venus-1.5-30B~\cite{team2026ui}. Variants are processed sequentially while carrying a single policy across the family. Construction starts from each method's own exploration and never exposes the original task. Only model-judge feedback is available during construction and benchmark evaluators are reserved for final evaluation. \frameworkabbr{} performs up to three refinement iterations with up to three judged rollouts per variant. $J_\text{task}$ determines rollout success, while $J_\text{loc}$ provides failure localization for healing. This gated setup was adapted due to the sequential aspect of the experiment meant that lenient judges are more appropriate over strict judges due to the chained dependency.

\textbf{AutoRPA adaptation.} We adapt the AutoRPA reference implementation~\citep{chen2026autorpa} to computer-use environments while preserving its exploration, reflection, breakpoint continuation, newest-first verification, and rollback logic. We remove inputs unavailable from a normal user request---the task template, named parameter schema, and benchmark reward. Both methods instead read the current request at execution time, use $J_\text{task}$ in place of reward, and share the same visual grounding model.

\textbf{Evaluation.} After construction on all variants, the policy is frozen and evaluated three times from fresh environments on the original task. If $c$ of the three runs succeed, the family contributes $\binom{c}{k}/\binom{3}{k}$ to \passk{k}; results are averaged across families, with construction failures assigned zero.

\subsection{Judge analysis details}
\label{sec:appendix:judge_analysis}

\textbf{Annotation.} We annotate 19 OSWorld tasks whose initial code policy failed according to $J_\text{task}$, one rollout per task (\Cref{tab:appendix:judge_cases}). We leave out tasks that fail because a Python library is missing in the guest; these are the most common failures, and the step-level judge cannot see them (see the first limitation). One annotator labels each rollout. The annotator sees the task, the benchmark evaluator, the code policy, one screenshot per step, and the output of every shell command and script. For each rollout, the annotator records five labels: 1) whether the rollout really failed; 2) the divergence step; 3) the first visible divergence, i.e., the first step at which a person who sees only the screenshots can tell that the rollout went wrong, or none if the failure never shows on screen; 4) the type of failure; 5) the cause, in one or two sentences. The annotator then sees the output of the step-level judge and rates it: whether $t^\ast$ is correct, adjacent (within the same sub-goal), or wrong; whether the narrative names the real cause; whether the suggested focus would steer the healer to the right fix; and whether the observer described the divergence step correctly.

\textbf{Ground truth.} The annotator judges success against the task instruction and uses the benchmark evaluator as a reference. The two disagree on one rollout, in which the plan wrote network speeds from a command-line tool instead of copying them from the web page that the task names. The evaluator only checks that the file names the three metrics, so it passes the rollout, but the annotator counts it as a failure.

\textbf{Task completion judge.} $J_\text{task}$ is correct on 16 of the 19 tasks. The annotator judges the other 3 tasks as successful, so $J_\text{task}$ gave false alarms on them. In two of the false alarms, $J_\text{task}$ read outdated values from saved files; in the third, it read a default value written by a web tool as a violation of the instruction.

\textbf{Step-level judge.} We compare $t^\ast$ with the first visible divergence on the 16 confirmed failures. In 6 of them, the failure happens only inside a script and is never visible to the step-level judge; $t^\ast$ is wrong in all 6. In the other 10, the failure becomes visible on screen at some step. Against the first visible divergence, $t^\ast$ is correct in 7 of these 10, adjacent in 2, and wrong in 1. The narrative names the real cause in 4 of the 10, names it in part in 4, and is wrong in 2. The suggested focus would steer the healer to the right fix in 7 of the 10.

\textbf{Errors in visible failures.} Of the 10 visible failures, $t^\ast$ misses the first visible divergence in 3. In one, the observer reported no change although a menu failed to open, so $t^\ast$ points four steps later, where the menu closes. In another, the plan fell one step behind; the judge named the step next to the divergence in its suspect chain but reported a later step as $t^\ast$. In the third, the judge matched a condition check to a click in the code, because condition checks write no step-log entry, and placed $t^\ast$ three steps before the failure shows on screen.

\textbf{Limitations.} We find four limitations.

First, the step-level judge receives no return values from the primitives. It sees the code policy and the screenshots, but not the output of shell commands and scripts or the answers of condition checks, and condition checks appear as screen transitions with no action. This limitation also hides the most common failure, a missing library, because the error appears only in the script output; our sample leaves these failures out, so the counts above are more favorable to the step-level judge.

Second, when a click misses or registers late, the code policy falls one step behind, and each later action hits the wrong target. The step-level judge then places $t^\ast$ at the later symptom instead of the step that the healer needs to fix. For instance, a short wait causes the next click to fire before the target has loaded.

Third, neither judge recognizes an infeasible task. Both infeasible tasks in the sample ask for something that does not exist: a ``Blue'' theme in GIMP and a photo in a folder that has none. Both judges suggest how to complete the task instead of reporting that it is infeasible.

Fourth, $J_\text{task}$ trusts saved files and the output of its own tools more than the running application. Saved files can lag the state of the application, and tool output can contain text that is not in the file.

\begin{table}[h]
    \centering
    \scriptsize
    \caption{The 19 annotated rollouts. Step: true divergence step. Visible: first visible divergence (--, never). $t^\ast$ rating against the true step and the first visible divergence: C correct, A adjacent, W wrong.}
    \label{tab:appendix:judge_cases}
    \begin{tabular}{llllllp{5.6cm}}
        \toprule
        ID & App & Step & Visible & $t^\ast$ & Rating & Cause \\
        \midrule
        R01 & Chrome & 27 & 27 & 27 & C / C & Site is sold out; plan should report the task as infeasible \\
        R02 & Chrome & 3 & 22 & 22 & W / C & \texttt{date -d monday} returns today, not next Monday \\
        R03 & GIMP & 9 & 9 & 13 & A / A & Click on Mode does not open the menu; next click reuses its point \\
        R04 & GIMP & 14 & 18 & 15 & A / W & Condition check returns empty; Export skipped; Replace dialog left open \\
        R05 & GIMP & 6 & 6 & 9 & A / A & Wait too short; click misses; plan falls one step behind \\
        R06 & GIMP & 2 & 5 & 5 & W / C & Infeasible: plan writes a ``Blue'' theme that does not exist \\
        R07 & Calc & 7 & 9 & 9 & A / C & Script turns numbers into text; formulas show \texttt{\#VALUE!} \\
        R08 & Impress & 8 & 8 & 8 & C / C & Click reopens the background list; Pattern set; hex typed into slide \\
        R09 & Multi & 2 & -- & 0 & W / -- & Regex bug; script appends headings instead of filling placeholders \\
        R10 & Multi & 9 & -- & 15 & W / -- & Values from a command-line tool, not the web page \\
        R11 & Multi & 5 & -- & 9 & W / -- & Invoice month taken from the email send date \\
        R12 & Multi & -- & -- & 8 & -- & Success; $J_\text{task}$ reads a tool default as a violation \\
        R13 & Multi & 5 & 5 & 5 & C / C & Drag and drop does not move the sheet tabs \\
        R14 & Multi & 2 & -- & 0 & W / -- & Conversion to xlsx fails after the plan kills the open Calc \\
        R15 & Multi & 4 & -- & 1 & W / -- & Unlabeled column; script raises an error and writes nothing \\
        R16 & OS & -- & -- & 4 & -- & Success; $J_\text{task}$ reads an old dconf value \\
        R17 & Thunderbird & -- & -- & 12 & -- & Success; $J_\text{task}$ reads the mail file before the star is saved \\
        R18 & VS Code & 2 & 7 & 7 & W / C & Infeasible: plan uses a server screenshot as the photo \\
        R23 & Impress & 4 & -- & 8 & W / -- & Script underlines the title as well as the body \\
        \bottomrule
    \end{tabular}
\end{table}

\subsection{Precondition checks}
\label{sec:appendix:precondition}

\begin{figure}[t]
\centering
\includegraphics[width=0.5\linewidth]{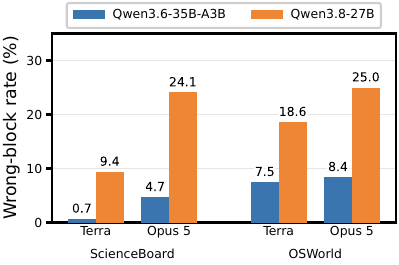}
\caption{Pre-action verifier error rates at $\theta = 0.78$. Rollout-level wrong-block rate on correct policies, by environment, base model used for initial trajectory seed, and verifier model.}
\label{fig:precondition_verifier}
\end{figure}
The verifier receives a natural-language explanation of each primitive alongside its code. At deployment, the runtime takes this explanation from the comments of the code policy: a trailing comment on the statement that contains the call, otherwise the block of comment lines directly above that statement, otherwise the first line of the docstring of the enclosing function, and otherwise a literal restatement of the call (e.g., ``press ctrl+q'', ``run a shell command on the VM''). The final code policies produced by policy iteration carry few such comments: on OSWorld with the Terra converter, 99.9\% of gated primitives fall back to the literal restatement. A literal restatement gives the verifier no reason for steps that are correct within the plan's strategy but look destructive in isolation, such as closing an application so that a later step can edit its file on disk. We therefore inject one comment per gated primitive that states the intent of the step within the overall code policy, using the following procedure.

\textbf{Targets.} We parse each plan and find every call to a gated primitive (\texttt{click}, \texttt{type}, \texttt{hotkey}, \texttt{open}, \texttt{scroll}, \texttt{drag\_and\_drop}, \texttt{set\_cell\_values}, shell and Python execution, etc.). We group calls by the statement that the runtime uses to look up the explanation, and we leave a statement unchanged if it already has a trailing comment or a comment on the line directly above it, so that existing comments written during refinement are preserved.

\textbf{Intent generation.} For each plan we make one call to Gemini 3.8 Flash (low thinking level) with the task instruction, the plan with line numbers, and the target statements, and the model returns one comment per target as JSON. Each comment starts with the action and its target in the words of the code, then explains why the plan takes the step at that point in its strategy. For a step that could look wrong out of context, e.g., closing or killing an application, saving, or running shell code instead of using the GUI, the comment says why the step is needed, and for a close, kill, or save it names the later step that depends on it. A step inside a retry branch is described as retrying an attempt that the preceding check found did not take effect. The comment must not rename or correct the target, claim facts the plan does not establish (e.g., that a file was already saved), or merely restate the call, and it fits on one line of at most 220 characters.

\textbf{Insertion and verification.} The model writes only comments, and a script inserts each one at a fixed position: on the line above the statement for shell and Python execution, whose code argument is usually long or spans several lines, and at the end of the line that ends the call for all other primitives. A call in the condition of an \texttt{if} or \texttt{while} statement gets its comment at the end of the header line, because the runtime's lookup treats the whole compound statement as the call's statement and would otherwise find a comment from the body first. We accept the annotated plan only if its abstract syntax tree is unchanged and the runtime's lookup returns the new comment for every annotated call; otherwise we call the model again, up to four attempts in total, and keep the original plan if none passes. We ran this procedure on the final code policies of OSWorld with the Terra converter (348 plans) and of ScienceBoard with the Opus~5 and Terra converters (159 and 167 plans), and it annotated 673 of the 674 plans with 4{,}321 comments; the remaining plan does not parse as Python.

\textbf{Verifier inputs.} All runs send a fixed note stating that the machine runs Ubuntu with GNOME, where Ctrl- and Alt-based shortcuts are native, because the task instructions given to the verifier do not name the operating system and the rubric blocks shortcuts meant for the wrong operating system. We also evaluate two optional additions to the production rubric. The \emph{exemptions} address correct plans, whose close, kill, save, and retry steps the production rubric reads as destructive or already done. They state that the following are not by themselves reasons to block: opening or focusing an application, window, or file that is already open; closing or killing an application when the explanation gives a task-related reason, unless the screen shows unsaved work that the step would discard; saving the open document; and repeating a step that the explanation says is a retry, or re-navigating to the page already shown. The \emph{blocked targets} add two reasons to block that the production rubric missed on perturbed initial states: a shortcut or typed text meant for a specific application while that application is closed, minimized, not in front, or has no document loaded; and any input aimed at a window that a dialog, popup, or menu covers, unless the input is aimed at the dialog itself. With blocked targets, the rubric's allowance for targets not visible in the screenshot no longer covers these two cases. We report results with and without these additions, since both change the rubric used in production.

\subsection{OSWorld-V2 extension}
\label{sec:appendix:osworld-v2}

This section records the details of the OSWorld-V2 experiment (\Cref{sec:results:longer-tasks}) that differ from the OSWorld-Verified and ScienceBoard setup, together with the infrastructure events that affected the runs.

\textbf{Benchmark version and task set.} We use the June 2026 release of OSWorld-V2 (\texttt{v2026.06.24}), with task definitions, evaluators, and assets unmodified. We report 107 of the 108 tasks. We exclude task 041 because its instruction hard-codes the benchmark authors' GitLab instance, while setup and grading use the locally deployed one, so no run can be scored as intended. Task 093 is included as a zero for our method, because the converter did not produce a program for it.

\textbf{Website hosting.} 39 of the tasks use web applications that were originally hosted by the benchmark authors. During our experiments, the authors stopped serving the \texttt{v2026.06.24} version of these applications and replaced it with a newer version at the same address. The switch happened between 2026-08-25 and 2026-09-10.

\textbf{Base agent.} The base agent is a custom harness driving Claude Opus~5 with maximum adaptive thinking and a budget of 500 steps per task, the budget that the benchmark protocol specifies for Claude models. Its seed trajectory for each task comes from run 1.

\textbf{Policy construction.} The loop follows the refinement algorithm with the settings in \Cref{tab:appendix:osworld-v2-params}. It differs from the OSWorld and ScienceBoard runs in four ways. First, the converter is an agent (Claude Opus~5) that rewrites the seed trajectory into a program, replacing pixel coordinates with element descriptions and deriving task values at run time instead of copying them from the trajectory. Second, the continuation agent is the custom harness running DeepSeek-V4.1-Flash, with 500 steps and a three-hour wall-clock cap, instead of the base model. Third, the step-level judge receives each rollout as a single transition: the screenshots before and after execution, and the program's output, capped at its last 50 log entries. It therefore returns a verdict and a failure mode but cannot localize the failure to a step. Fourth, the healer also rewrites the policy after a failure in the last iteration, and the rewritten policy is returned without being judged. As in the other benchmarks, the loop skips the judges and the continuation for rollouts that end in a code execution error.

\begin{table}[h]
\caption{Settings of the OSWorld-V2 construction runs.}
\label{tab:appendix:osworld-v2-params}
\centering
\small
\begin{tabular}{ll}
\toprule
Setting & Value \\
\midrule
Iterations $N$ / rollouts per iteration $K$ / final trials & 5 / 3 / 3 \\
Converter $\Phi_\text{conv}$ and healer $\Phi_\text{heal}$ & Claude Opus~5 \\
Continuation agent $\mu$ & harness + DeepSeek-V4.1-Flash, 500 steps, 3\,h cap \\
Task-completion judge $J_\text{task}$ & Claude Sonnet~4.6, 60 tool calls, 900\,s \\
Observer / localizer & Gemini 2.5 Flash / Gemini 2.5 Pro \\
Screenshots per heal & 5, evenly spaced \\
Construction budget & \$60 per task (never reached; max \$27.66) \\
Task wall-clock limit & 10\,h (exceeding it counts as an infrastructure failure) \\
\bottomrule
\end{tabular}
\end{table}

\textbf{Infrastructure reruns.} We rerun a task only when an infrastructure fault affected its result, never because of a low score, and every completed rerun replaces the original score, including when the score goes down. Task instructions, rubrics, and score weights are unchanged. We ran 13 reruns on tasks with infra failures (004, 005, 010, 018, 021, 024, 025, 058, 061, 073, 089, 092, 100). Of these, four scores went up (004, 025, 092, 100), one went down (061), and eight were unchanged. Tasks whose reruns are still running keep their original scores.

\textbf{Cost.} Construction cost \$8.2 per task on average in LLM calls (Opus~5 58\%, Sonnet~4.6 30\%, DeepSeek 9\%, Gemini 3\%) and a median of 2.9 VM-hours per task. Counting infrastructure reruns, total LLM spend was $1.8\times$ that amount. Grounding calls and LLM calls inside official evaluators are not metered. We estimate script costs during execution due to LLM calls to be \$0.03–\$0.08 on average.

\section{Constructing Policies from Unsuccessful Trajectories}

\label{sec:appendix:failed_seeds}
To examine whether NSPI can build a working policy from an unsuccessful trajectory. We split the runs by whether the selected seed trajectory $\tau_\mu$ succeeds under the benchmark evaluation script. A failed seed is from tasks on which the base agent fails all three attempts. The resulting split represents how NSPI behaves when it starts from a successful versus an unsuccessful demonstration of the task.

We find that NSPI can construct reliable policies from unsuccessful trajectories to a substantial degree as shown in Figure~\ref{fig:results:initial_trajectory}. Across the 356 policies built from a failed seed, 62 succeed in all three final runs under the benchmark evaluator, reflecting policies that are consistently correct rather than occasionally lucky. On tasks with a successful seed, the final plan also improves over the base agent in three of the four settings, indicating that NSPI generally adds reliability even when a correct demonstration is already available.

Recovering from a failed seed is more costly than refining a successful one. Failed seeds cost 18--31\% more and take 17--38\% longer to reach a policy than successful seeds, and in three of four settings they receive a larger share of judge rejections during policy iteration . This is consistent with the intuition that a failed trajectory contains more incorrect or missing steps that must be diagnosed and repaired before the resulting policy passes the judges.

\end{document}